\documentclass[review,12pt]{elsarticle}

\usepackage{lineno}        % Line numbers (required for review)
\usepackage{hyperref}      % Hyperlinks
\usepackage{amsmath}       % Math
\usepackage{amssymb}
\usepackage{graphicx}      % Figures
\usepackage{booktabs}      % Professional tables
\usepackage{natbib}        % Citations (author-year style)
\usepackage{xcolor}
\usepackage{caption}
\usepackage{subcaption}    % Sub-figures
\usepackage{float}         % For [H] float placement
\usepackage{geometry}
\usepackage{enumitem}
\usepackage{longtable}
\definecolor{revisionblue}{RGB}{0,70,180}
\definecolor{revisionred}{RGB}{190,0,0}

\newcommand{\revremove}[1]{{\color{revisionred}#1}}

\modulolinenumbers[5]

\journal{Journal of Results in Engineering}

\begin{document}

\begin{frontmatter}

%% Title
\title{
World Models for Ocean Engineering: A Critical Review of Maritime Autonomy, Adaptive Latent Prediction, Digital Twins, and Biodiversity Monitoring}
%\title{WMs: Current Advances, Challenges, and Future Perspectives In Marine Unmanned Surface Vehicles}

%% Authors
\author[aff1,aff4]{Basma Dokkar}
\ead{basma.dokkar@unisalento.it}

\author[aff2]{Esteban W. G. Clua}
\ead{esteban@ic.uff.br}

\author[aff3]{Luiz M. G. Gon\c{c}alves}
\ead{lmarcos@dca.ufrn.br}

\author[aff4]{Paolo Spagnolo}
\ead{paolo.spagnolo@cnr.it}

\author[aff1,aff4]{Cosimo Distante\corref{cor1}}
\ead{cosimo.distante@cnr.it}

%% Affiliations
\affiliation[aff1]{organization={Department of Innovation Engineering, University of Salento},
             addressline={via Monteroni},
             city={Lecce},
             postcode={73100},
             country={Italy}}

\affiliation[aff2]{organization={Universidade Federal Fluminense},
             city={Niteroi},
             country={Brazil}}

\affiliation[aff3]{organization={Universidade Federal do Rio Grande do Norte},
             city={Natal},
             country={Brazil}}
\affiliation[aff4]{organization={Institute of Applied Sciences and Intelligent Systems, CNR},
             addressline={via Monteroni},
             city={Lecce},
             postcode={73100},
             country={Italy}}

%% Corresponding author
\cortext[cor1]{Corresponding author.}

%%----------------------------------------------
%%  ABSTRACT
%%----------------------------------------------
\begin{abstract}
% \begin{revremoveblock}
% We provide a comprehensive overview of WMs, covering the theoretical
% foundations, recent advances, and open challenges in the field. We systematically
% surveyed the literature from [start year] to [end year], identifying [N] relevant
% studies from major databases including Scopus and Web of Science. After analysis, we emphasize Joint Embedding Predictive Architectures (JEPA) as the leading paradigm and extend the review to other paradigms and main applications.
% \end{revremoveblock}

Marine robots face partial observability, non-stationary dynamics, and sparse communication, motivating WMs that learn compact predictive states for planning. This review develops a functional taxonomy of WMs for ocean engineering, focusing on Joint-Embedding Predictive Architectures (JEPAs) and AdaJEPA's test-time adaptation within model predictive control, mapping them to navigation, adaptive sampling, digital twins, and biodiversity monitoring, and separating demonstrated evidence from prospective applications. The synthesis identifies key Gaps:—benchmarks, uncertainty calibration, safe adaptation—and proposes a research agenda for physics-informed, deployment-adaptive marine WMs validated through simulation and field missions.

\end{abstract}

%%----------------------------------------------
%%  KEYWORDS
%%----------------------------------------------

\end{frontmatter}

%% Line numbers
%\linenumbers

%%==============================================
%% 1. INTRODUCTION
%%==============================================
\section{Introduction}
\label{sec:introduction}
% 1) Introduce ML/AI, the challenges of classical static AI and the need as well as the natural resulting of dynamic AI (WMs)
The world's oceans are fundamental to climate regulation, biodiversity, transportation, and sustainability, yet remain among the least-observed domains because of their vast extent, dynamic nature, and harsh conditions.
Recent advances in marine robotics have enabled autonomous platforms—USVs, AUVs, sailboats, gliders—equipped with multimodal sensors generating unprecedented heterogeneous data. 

However, current autonomy frameworks remain largely reactive, relying on waypoint-following and task-specific perception disconnected from broader environmental understanding—inadequate for the uncertain, partially observable marine environments where decision-making requires anticipating future conditions over extended horizons.

To address these limitations, the field of Artificial Intelligence is increasingly moving toward the concept of WMs. This concept is not novel; it dates back to state-space representations in control theory. In a general sense, WMs can be understood as internal representations (data structures) working as "simulators" that allow software agents to envision future states before executing actions~\cite{Ha2018}. Rather than reacting solely to current observations, agents equipped with WMs can plan through imagination, evaluate alternative courses of action, and operate more effectively under uncertainty. 

%%%% new added
The emergence of Foundation Models (FM) and Self-Supervised Learning (SSL) has significantly accelerated progress in this direction. In particular, Joint Embedding Predictive Architectures (JEPAs) have recently introduced a paradigm shift from generative reconstruction toward predictive representation learning. Rather than reproducing sensory observations at the pixel level, JEPAs learn latent representations that capture the environment's essential structure and dynamics, enabling more efficient planning and decision-making in complex physical systems.

While WMs have attracted considerable attention in autonomous driving, robotics, and embodied AI, their potential in marine systems remains largely unexplored—a significant gap, since marine environments naturally exhibit many of the challenges WMs are designed to address, including partial observability, sparse communications, sensor degradation, and strong environmental uncertainty.

Beyond navigation and control, WMs open opportunities for ecosystem-scale intelligence: by integrating observations from autonomous vehicles, remote sensing, and acoustic monitoring, future WMs could serve as the predictive core of marine digital twins forecasting biodiversity dynamics and supporting adaptive conservation, with autonomous sailboats and USVs becoming active sensing agents in continuously evolving ecosystem representations.

For this review, we treat a model as a \emph{WM} only when it represents environmental state, predicts temporal evolution, and exposes those predictions to forecasting, planning, or control. A representation learner such as an image- or audio-only JEPA therefore becomes a WM only when coupled to temporal dynamics and, for embodied autonomy, to actions or interventions. This operational definition prevents conflating self-supervised representation learning, generative simulation, and action-conditioned WMing.

\begin{figure}
    \centering
    \includegraphics[width=5in, height=3in]{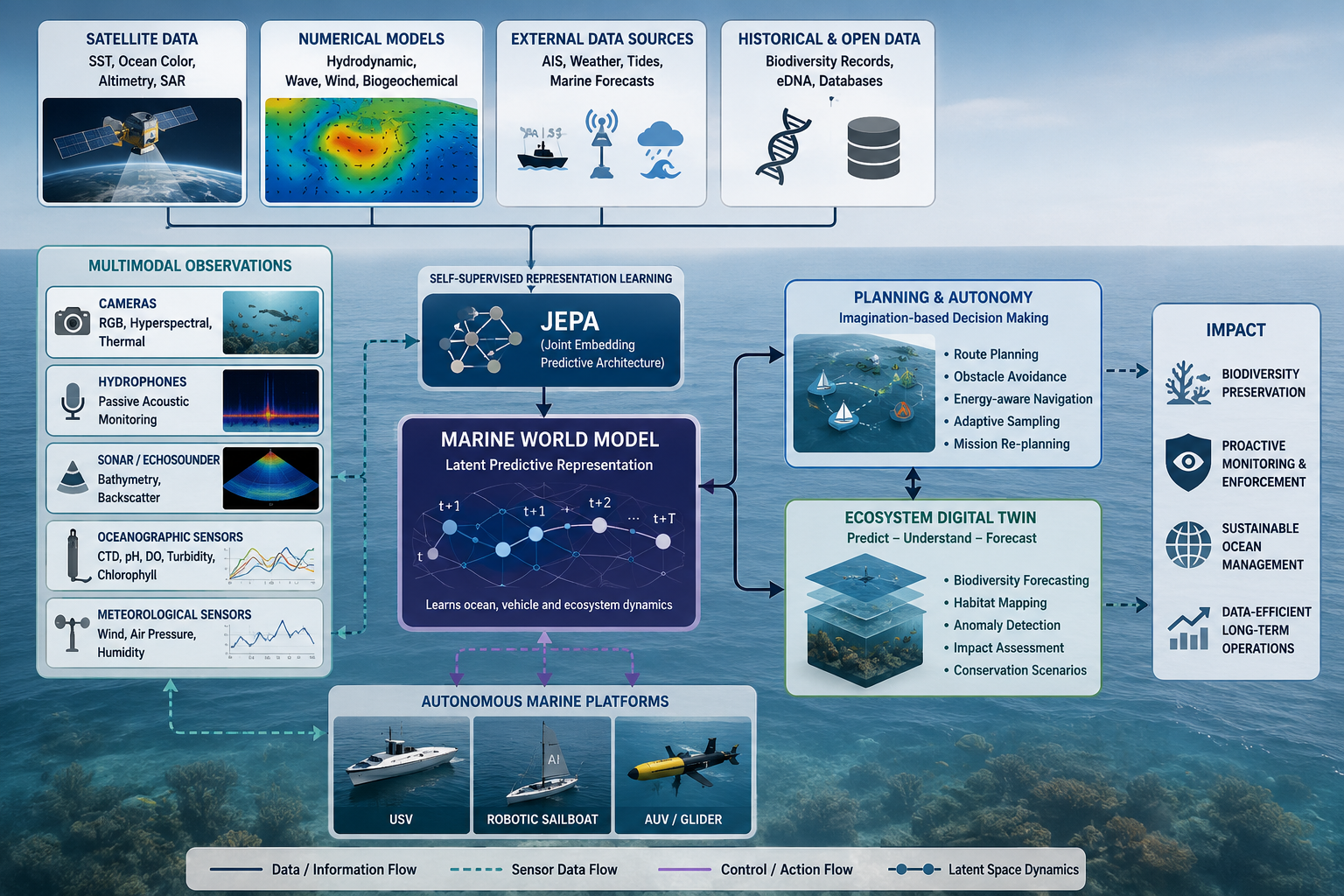}
    \caption{Conceptual framework of Marine WMs for Ocean Intelligence}
    \label{fig:marine_wm}
\end{figure}

Figure~\ref{fig:marine_wm} summarises this review's vision: future autonomous ocean systems will rely on predictive latent representations integrating heterogeneous observations into a unified Marine WM, jointly supporting robotic decision-making and ecosystem understanding rather than treating perception, navigation, and monitoring as separate tasks.
As shown, information originates from complementary sources: remote sensing provides large-scale oceanographic observations, hydrodynamic models contribute physical priors on currents and forcing, and marine robotic platforms acquire local observations via hydrophones, cameras, sonars, and oceanographic sensors.

These heterogeneous data streams are projected into a shared latent representation through a Joint Embedding Predictive Architecture (JEPA)~\cite{lecun2022path}. Rather than reconstructing observations at the pixel level, the JEPA learns predictive embeddings that capture the underlying dynamics of both the marine environment and the robotic platform. This latent representation forms the core of the Marine WM, enabling the prediction of future environmental states and the evaluation of alternative action sequences.

The Marine WM supports two complementary objectives: enabling autonomous planning for marine robots (adaptive sampling, collision avoidance, route optimisation, energy-aware navigation) and providing the predictive engine of ecosystem digital twins that assimilate multimodal observations to forecast biodiversity dynamics and environmental anomalies. In this view, autonomous sailboats and USVs are active agents contributing to the continuous construction of marine ecosystem representations, not merely sensing platforms.

Motivated by these developments, this paper critically reviews modern WMs, with particular emphasis on predictive latent architectures and their potential role in ocean autonomy. We analyse the evolution from generative approaches to recent JEPA-based frameworks, discuss their application to physical AI systems, and identify emerging opportunities for autonomous marine robotics, biodiversity monitoring, adaptive environmental sampling, and ecosystem digital twins.
%%%% end on new text added

% 2) Introduce WMs
% 3) Explain the paper structure 
% 4) Contribution
The main contributions of this review are:
% \begin{revremoveblock}
% \begin{itemize}
% %%%%new added items 
%     \item To provide a comprehensive overview of modern WM architectures, including generative, predictive, and JEPA-based approaches.
%     \item To establish a conceptual framework linking WMs with ocean autonomy, marine robotics, and long-endurance autonomous platforms.
%     \item To establish a conceptual framework linking WMs with ocean autonomy, marine robotics, and long-endurance autonomous platforms.
%     \item To examine the potential of WMs for marine ecosystem intelligence, biodiversity monitoring, and environmental digital twins.
% %%%% end of new added items     
%     \item to discover and show case the state-of-the-art of scientific literature about ML-WM;
%     \item pointing out the main contributions to Physical AI systems such as Unmanned Surface Vehicles;
%     \item an analysis of the works done in this context, pointing future research directions;
% \end{itemize}
% \end{revremoveblock}

\begin{itemize}
    \item A functional taxonomy that separates observation reconstruction, latent prediction, temporal dynamics, action conditioning, planning, and deployment-time adaptation.
    \item A critical synthesis of marine evidence across autonomous navigation, adaptive ocean observation, digital twins, and biodiversity-monitoring missions.
    \item An analysis of AdaJEPA as a deployment-adaptive WM, including its closed-loop mechanism, demonstrated benefits, computational assumptions, and limits of transfer to marine platforms.
    \item An evidence-aware research agenda covering realistic marine distribution shifts, uncertainty calibration, safe adaptation, multi-rate sensor assimilation, embedded computation, and field validation.
\end{itemize}

% \begin{revremoveblock}
% The remainder of this paper is organized as follows:  The review is organized as follows: Section~\ref{sec:background} presents the
% background and scope; Sections~\ref{sec:methods}--\ref{sec:results} detail
% methodological trends and key findings; Section~\ref{sec:discussion} discusses
% implications and limitations; and Section~\ref{sec:conclusion} concludes with recommendations for future research.
% \end{revremoveblock}

The review first motivates WMs from marine operational requirements and then establishes the architectural foundations needed to distinguish generative, predictive-latent, and hybrid approaches. Subsequent sections examine JEPAs and deployment-time adaptation, define the review protocol and evidence categories, synthesise maritime and biodiversity applications, and conclude with evaluation priorities and a research agenda for ocean engineering.

\section{Why Marine Environments Need WMs}

Marine environments pose significant challenges for autonomous systems due to variable conditions, sparse observations, uncertain dynamics, limited communication, and complex physical, biological, and human interactions. These challenges motivate predictive approaches such as WMs, which allow autonomous agents to anticipate future states, evaluate possible actions, and make decisions under uncertainty. Such capabilities are particularly important for long-duration missions, adaptive monitoring, and large-scale ecosystem observation.
\subsection{Partial Observability and Sparse Measurements}
Marine environments are highly partially observable, requiring autonomous systems to infer hidden states from sparse and heterogeneous sensor data. This challenge is even greater in ecological monitoring, where biodiversity must often be inferred indirectly through sources such as acoustic recordings and eDNA~\cite{stephenson2020technological}. WMs provide a suitable framework for integrating these diverse observations into predictive latent representations of marine ecosystems.
\subsection{Environmental Uncertainty and Multi-Scale Dynamics}
Marine systems are governed by complex, nonlinear interactions across multiple spatial and temporal scales. Long-endurance autonomous missions therefore require anticipating future environmental conditions, such as wind and energy availability, rather than relying only on short-term predictions. WMs address this challenge by learning predictive environmental representations that enable agents to simulate possible futures and assess the consequences of actions before deployment.
\subsection{Communication Constraints and Autonomous Decision Making}
Marine robotic platforms often operate with limited or intermittent communication—satellite links are costly and low-bandwidth, while underwater communication remains highly constrained~\cite{ning2026event}—so they require onboard intelligence rather than continuous human or cloud support. WMs enable this through local reasoning: predictive latent representations let agents evaluate future states, estimate risks, and adapt behaviour without external guidance, particularly for sailboats, gliders, and long-endurance USVs operating far from shore.
\subsection{Adaptive Ocean Observation and Ecosystem Monitoring}
The global ocean remains poorly observed because conventional monitoring is costly and has limited coverage. Autonomous vehicle fleets provide a scalable alternative, while WMs can enable adaptive monitoring by identifying uncertain regions and guiding platforms toward the most informative observations~\cite{li2026traditional}. This is particularly valuable for biodiversity monitoring, where autonomous sailboats can use hydrophones, imaging, and eDNA sampling to detect ecological hotspots and support conservation efforts.
\subsection{Towards Marine WMs}
Together, these challenges—partial observability, environmental uncertainty, communication constraints, and sparse ecological observations—position marine environments as a natural domain for WMs. We therefore introduce \emph{Marine WMs}: predictive latent representations jointly capturing ocean dynamics, ecosystem processes, vehicle behaviour, and anthropogenic influences, linking autonomous robotics, digital twins, and biodiversity forecasting (Figure~\ref{fig:marine_wm})—motivating the review's focus on WMs' evolution and role in ocean intelligence.

\begin{longtable}[!h]{p{2.5cm} p{7.5cm} p{7.5cm}}
\caption{Key challenges of autonomous marine systems and the potential advantages offered by WMs.}
\label{tab:marine_challenges} \\

\toprule
\textbf{Marine Challenge} &
\textbf{Limitations of Conventional Approaches} &
\textbf{WM Advantage} \\
\midrule
\endfirsthead

\midrule
\endhead

\bottomrule
\endlastfoot

Partial Observability &
Environmental states are only partially measured, and many relevant variables remain hidden. &
Latent state representations enable inference of unobserved environmental and ecological processes. \\

Sparse Measurements &
Observations are spatially and temporally fragmented across heterogeneous sensing platforms. &
Multimodal data fusion combines information from satellites, sensors, robotic platforms, and environmental models. \\

Environmental Uncertainty &
Reactive control strategies struggle to anticipate evolving ocean conditions. &
Predictive latent dynamics support forecasting and proactive decision-making. \\

Communication Constraints &
Continuous human supervision is often infeasible due to limited bandwidth and intermittent connectivity. &
Onboard predictive reasoning enables autonomous operation under limited communication. \\

Biodiversity Monitoring &
Ecological observations are sparse, noisy, and collected through heterogeneous sensing modalities. &
Shared ecosystem representations facilitate biodiversity forecasting and ecological state estimation. \\

Long-Duration Missions &
Traditional planning methods have limited capability to reason over extended temporal horizons. &
Imagination-based planning enables long-horizon prediction and adaptive mission execution. \\

\end{longtable}

Table~\ref{tab:marine_challenges} summarises the principal challenges faced by autonomous marine systems and highlights how WMs provide a unified framework for addressing many of the limitations of conventional perception-driven autonomy architectures.

%\subsection{Overview of Previous Surveys}

The rapid expansion of Foundation Models has driven a surge in survey literature, though existing works focus on specific silos. General \textit{WMs} surveys mainly address RL for video games or simple robotic manipulation~\cite{ha2018world}. In \textit{Surgical Robotics}, prior reviews focus on traditional imitation learning and physics-based simulators, predating the shift toward video-based WMs~\cite{shahid2026applications}. Similarly, \textit{Earth Observation} surveys frequently discuss contrastive learning or MAEs but rarely address action-conditioning for physical autonomy~\cite{bagheri2026multisensor}, while \textit{Wireless Communications} has explored AI for resource management, but the intersection of latent dynamics and "imagination-based" scheduling remains nascent.

% Other work instead traces the evolution of WMs, their emerging functionalities and modalities, and primary applications. Ding et al.~\cite{ding2025understanding} provide a comprehensive review organized around two functional categories—\textit{understanding the world and predicting the future}—and three key applications: \textit{autonomous driving, robotics, and social simulacra}. While it captures conceptual and practical advances and challenges well, it lacks a formalized search strategy, making inclusion/exclusion criteria unclear and the review more narrative than systematic.

% Kawaharazuka et al.~\cite{kawaharazuka2025vision} review Vision Language Action Models (VLAs), positioning WMs as a VLA sub-category that interacts with vision, language, and action to predict future states for \textit{planning} and \textit{reasoning}, and to guide action via an embedded inverse dynamics model generating motor output from visual observations or their latent embeddings for \textit{control}. The study also examines WM types paradigmatically, covering diverse applications, design choices, and usability for simulation and data generation—though its broad, modality-oriented scope blurs distinctions between different WM architectures..
{\renewcommand{\arraystretch}{1.3}
\begin{longtable}{lp{6.5cm}|lp{6.5cm}}
\caption{List of Acronyms and Abbreviations} \label{tab:acronyms} \\
\toprule
\textbf{Acronym} & \textbf{Full Expansion} & \textbf{Acronym} & \textbf{Full Expansion} \\
\midrule
\endfirsthead

\midrule
\endhead

\bottomrule
\endlastfoot

WM   & World Model & RSSM & Recurrent State-Space Model \\
SSL  & Self-Supervised Learning & JEPA & Joint Embedding Predictive Architecture \\
FM   & Foundation Models & USV  & Unmanned Surface Vehicle \\
WFM  & World Foundation Model & & \\
SSM  & Stochastic State-Space Model & & \\

\end{longtable}
}
%\caption{Summary of Previous Surveys}
%\centering
{\renewcommand{\arraystretch}{1.3}
\begin{longtable}{lp{12cm}}
\caption{Summary of Previous Surveys} \\
\toprule
\textbf{Reference} & \textbf{Findings and Identified Gaps} \\
\midrule
\endfirsthead

\midrule
\endhead

\bottomrule
\endlastfoot
Ding et al. ~\cite{ding2025understanding} (2025)&
\textbf{Taxonomy} WM functions: Predicting the future vs Understanding the World.
\textbf{Gaps:} The study is a general-purpose review which views WMs from a functional perspective, highlighting two categories: predicting the future and understanding the world. This dichotomy captures important local distinctions but leaves room for more unified global understanding of WMs, particularly for under-researched directions and applications like ocean intelligence. 
\\

Kawaharazuka et al.~\cite{kawaharazuka2025vision} (2025) &
\textbf{Taxonomy:} Vision-Language-Action Model types: Including WMs, Sensorimotor Models, and Affordance-based Models.
\textbf{Gaps:} The concept of WM is narrowed down to a design or mode of interaction with modalities, which overlooks the structural functionalism of WMs that is, to a certain extent, independent of the modality in question.\\

Zong et al.~\cite{zong2024self} (2024) &
\textbf{Taxonomy:} Multi-modal Self-Supervised Learning sub-paradigms, challenges, and solutions. 
\textbf{Gaps:} This work focuses on the paradigmatic aspects of self-supervised multi-modal and representation learning, without exploring WMs as a major advancement in the area. While emphasising multi-modal alignment, its cross-modality scope remains largely limited to alignable modalities rather than real physical-world modalities, including motor and sensory information.
\\
Ferdaus et al.~\cite{FERDAUS2026104598} (2026) & 
    \textbf{Taxonomy:} Evaluation of JEPAs for Autonomous Vehicles: Safety, security, and technical implementation.
    \textbf{Gaps:} This application-driven review focuses on Autonomous Vehicles, primarily comparing JEPAs' technical advantages and limitations against other state-of-the-art AV models. While it provides a solid JEPA background, it frames WMs as a JEPA subcategory—conceptually acceptable but JEPA-centric—leaving room to explore the topic from the WMs perspective instead, JEPAs' most promising fruit.
\\
\end{longtable}
}

% \subsection{Contributions of This Work}
% This survey is distinguished from prior literature by three primary contributions:

% \begin{enumerate}
%     \item \textbf{Cross-Domain Synthesis:} Unlike previous works that isolate robotics, telecommunications, and geospatial analysis, this paper identifies the \textbf{Joint Embedding Predictive Architecture (JEPA)} as a unifying framework for Physical AI across these diverse high-stakes industries.
%     \item \textbf{Predictive vs. Generative Focus:} We explicitly move beyond the "Generative AI" hype to analyze the "Predictive Latent" paradigm. We describe why predicting in embedding space is the superior solution for environments characterized by high noise (Marine), limited bandwidth (Wireless), and complex physics (Surgery).
%     \item \textbf{Action-Conditioned Analysis:} While most surveys treat vision as a passive task, our work focuses on \textbf{Vision-Language-Action (VLA)} integration. We evaluate how JEPA-based WMs allow robots to "reason" about the consequences of their movements, a topic that has not been comprehensively reviewed in the context of combined terrestrial, marine, and clinical applications.
% \end{enumerate}

%%==============================================
%% 2. BACKGROUND
%%==============================================
\section{The Philosophy of WMs}

The concept of a \textit{WM} rests on the hypothesis that autonomous agents achieve higher intelligence by maintaining an internal representation of their environment. Rather than reacting to instantaneous inputs, an agent with a WM can simulate future scenarios, "imagining" action consequences before physical execution—a foundation for \textit{Physical AI} enabling spatial reasoning and "common sense" understanding of physical laws. Historically, this relied on generative frameworks reconstructing observations at the pixel level. However, the field is shifting away from full-scale reconstruction—computationally prohibitive and prone to focusing on task-irrelevant details—toward a \textit{predictive} paradigm, where the goal is to predict underlying dynamics rather than replicate the world. This shift sets the stage for architectures like JEPA, prioritising essential latent structures over redundant visual generation.

\section{The Generative Approach}
\subsection{Generative WMs}
\textbf{Dreamer}, built on PlaNet~\cite{hafner2019learning}, forms the foundation of DreamerV1--V3~\cite{hafner2019dream,hafner2020mastering,hafner2023mastering}, introducing the Recurrent State-Space Model (RSSM): a VAE-based latent dynamics model combining a deterministic RNN ($h_t = f(h_{t-1}, s_{t-1}, a_{t-1})$) with a stochastic SSM ($s_t \sim p(s_t|h_t)$), minimizing KL divergence between prior and posterior states to force multi-step unrolling. DreamerV1 pairs RSSM with a CNN; DreamerV2 uses categorical latents, surpassing human-level Atari performance; DreamerV3 adds a \textit{symlog} transformation preventing training collapse, generalising to multitask robotic agents~\cite{wu2023daydreamer}. The \textbf{Cosmos World Foundation Model}~\cite{agarwal2025cosmos} instead treats the WM as a digital twin, predicting $\hat{x}_{t+1}$ via diffusion or autoregressive generation, but it fails when objects appear suddenly from below the frame. \textbf{NVIDIA OmniDreams}~\cite{basant2026nvidia} builds on pretrained Cosmos diffusion for closed-loop, action-conditioned generation using text, history frames, and a 3D abstract world state. \textbf{Cosmos 3}~\cite{agarwal2026cosmos} extends this as an omnimodal WM via a mixture-of-transformers architecture, with diffusion-based WMing attending to an autoregressive reasoning flow. Finally, \textbf{Qwen-RobotWorld}~\cite{qwenteam2026qwenrobotworld} is a language-conditioned embodied WM using a double Multi-modal Diffusion Transformer to decode language-action and visual-state streams from frozen VL (Qwen 2.5 VL) and VAE encoders.

Historically, WMs \cite{ha2018world} relied on generative processes to reconstruct observations. We discuss the limitations of this approach:
Generative WMs face three key limitations: \textbf{computational overhead} from the extreme cost of high-resolution pixel generation; the \textbf{task-irrelevant detail problem}, where generating full-dimensional scenes (e.g., every leaf on a tree) is unnecessary for downstream tasks and biases the model toward reconstruction rather than environmental understanding; and \textbf{hallucination}, where the generative objective inherently produces inconsistent predictions, undermining control over physical dynamics and realism.

Generative and predictive-latent approaches should not be seen as mutually exclusive, as many effective WMs combine compact latent dynamics with reconstructive, generative, or reward-prediction objectives. For ocean engineering, the key comparison is not whether a model generates pixels or observations, but which state variables it preserves, whether it predicts action-conditioned evolution, how it handles uncertainty, and whether its predictions support safe decisions under marine operational constraints.

\section{JEPA: The Predictive Shift}
% \begin{revremoveblock}
% As a "hot topic" in the current literature, the \textbf{Joint Embedding Predictive Architecture (JEPA)} represents a paradigm shift in SSL where the generative approach long served as the standard paradigm.
% JEPAs evolved from merely learning uni-modal representations to multi-modal WMing.
% \subsection{Predicting Latents over Pixels}
% Instead of reconstructing the input, JEPA predicts the \textbf{latent representation} of masked or future states. 
% \begin{itemize}
%     \item \textbf{Abstract Reasoning:} JEPA learns the "semantics" of a scene rather than its exact visual texture. 
%     \item \textbf{Sample Efficiency:} How self-supervised pre-training on diverse datasets (like AnySat's GeoPlex) leads to faster downstream finetuning.
% \end{itemize}
% \end{revremoveblock}

JEPA is a predictive representation-learning approach in which a predictor estimates target embeddings from context embeddings, rather than reconstructing every observable detail~\cite{lecun2022path}, reducing sensitivity to task-irrelevant variability. However, a JEPA is not automatically a WM—for maritime use, it must also model temporal evolution, incorporate actions or external forcing when decisions affect future state, and feed predictions to a controller, planner, or forecasting system. Accordingly, we analyse JEPA-based methods along independent functional axes: prediction target, temporal horizon, action conditioning, modality, planning interface, uncertainty treatment, and deployment mode, distinguishing models frozen after pretraining from those that adapt using operational observations.

\subsection{Architectural Evolution and Deployment Modes}

Image, video, audio, point-cloud, object-centric, and trajectory variants show that joint-embedding prediction can apply to several observation structures~\cite{assran2023self,bardes2023v,tuncay2025audio,saito2025point,nam2026causal}. Their relevance to WMing nevertheless differs. Image- and audio-level objectives primarily learn representations; video methods add temporal prediction; and action-conditioned approaches such as ACT-JEPA and PLDM connect latent dynamics to control or goal-conditioned planning~\cite{vujinovic2025act,sobal2026learning}. This distinction is important in marine systems, where recognising a scene or acoustic event is insufficient unless the model also anticipates how environmental forcing and vehicle actions change future state.
\begin{longtable}{p{2.0cm}p{7.0cm}p{8.0cm}}
\caption{Functional dimensions for comparing WMs in ocean engineering.} \label{tab:functional_taxonomy} \\
\toprule
\textbf{Dimension} & \textbf{Representative choices} & \textbf{Ocean-engineering relevance} \\
\midrule
\endfirsthead

\midrule
\endhead

\bottomrule
\endlastfoot

Prediction target & Pixels, tokens, objects, latent states & Determines which physical, ecological, and task-relevant information the model retains. \\
Temporal model & One-step, multi-step, hierarchical, stochastic & Controls useful forecast horizon and accumulation of model error. \\
Conditioning & Passive observation, action-conditioned, externally forced & Separates representation learning from models usable for vehicle control or intervention analysis. \\
Planning interface & Learned policy, gradient MPC, CEM, information-gain planning & Connects predicted futures to navigation, sampling, and mission objectives. \\
Deployment mode & Frozen, test-time adaptive, continual & Determines resilience to sensor, appearance, dynamics, and geographic shifts. \\
Evidence level & Adjacent-domain simulation, marine simulation, model-scale, field trial & Prevents transfer hypotheses from being presented as demonstrated marine capability. \\

\end{longtable}

\subsection{Deployment-Time Adaptation: AdaJEPA}
Most latent WMs remain frozen after offline training, even when prediction errors emerge during deployment. AdaJEPA instead inserts self-supervised adaptation into the closed loop of model predictive control (MPC)~\cite{wang2026adajepa}. At replanning step $t$, the agent plans with the current model, executes the first action chunk, observes the next transition, adds that transition to a recent buffer, updates a selected subset of encoder or predictor parameters, and replans with the adapted model. Using observation encoder $E_{\phi}$, action encoder $E_{\psi}$, latent predictor $f_{\theta}$, and recent transition buffer $\mathcal{B}_t$, the update can be summarized as
\begin{equation}
\mathcal{L}_{\mathrm{ada}}(\mathcal{B}_t)=
\frac{1}{|\mathcal{B}_t|}
\sum_{(o,a,o')\in\mathcal{B}_t}
\ell\!\left(f_{\theta}\!\left(E_{\phi}(o),E_{\psi}(a)\right),
\operatorname{sg}\!\left[E_{\phi}(o')\right]\right),
\qquad
\Omega \leftarrow \Omega-\eta\nabla_{\Omega}\mathcal{L}_{\mathrm{ada}},
\end{equation}
where $\Omega\subseteq\{\phi,\psi,\theta\}$ denotes the parameters permitted to adapt and $\operatorname{sg}$ is a stop-gradient operation.

We evaluated AdaJEPA on PushT and PointMaze under visual, geometry, dynamics, and layout shifts, with an adaptation overhead of $0.01$–$0.03$~s per step on an NVIDIA H200. These results establish proof of concept for adaptive latent planning but not marine readiness: experiments are simulated, short-horizon, and start pretrained, with no marine sensors or long-duration ecological processes evaluated. Maritime deployment involves analogous but distinct shifts—turbidity, biofouling, wave-induced motion—requiring uncertainty-gated updates, corrupted-transition rejection, rollback, and embedded-hardware evaluation. For biodiversity missions, delayed eDNA observations and seasonal change require slower multi-rate assimilation, not direct extrapolation from existing experiments.

\begin{figure}[!h]
    %\centering
    \includegraphics[width=4in, height=2.5in]{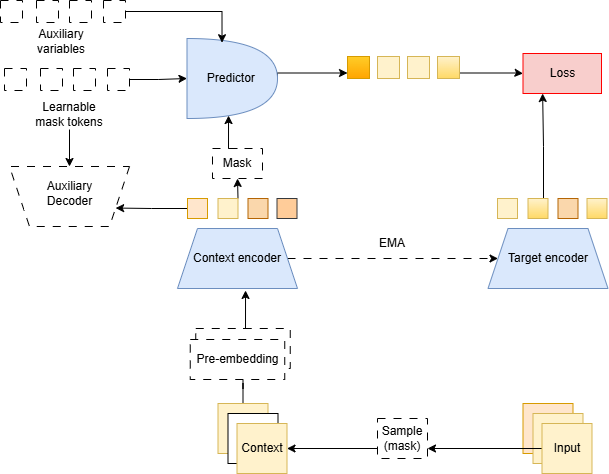}
    \caption{General framework of JEPAs. The main architecture includes a joint embedding (context encoder and target encoder), a predictor, and a loss. The architecture can be adapted to more complex data types through pre-embedding processes such as \textit{sequencing} and \textit{token embedding}. The predictor can benefit from auxiliary variables such as \textit{actions} and \textit{proprioceptive embeddings}.}
    \label{fig1}
\end{figure}

% \begin{longtable}{l|l|l|l|l|l}
% \caption{}
% \centering
% \\ JEPA & Column2 & Column3 & Column4 & Column5 & Column6 \\ \hline
% \end{longtable}

% \begin{revremoveblock}
% \section{Applications in Physical AI}
% We analyze how the JEPA-driven WM philosophy is applied across diverse, high-stakes environments:

% \subsection{Surgical Autonomy (Cosmos-Surg-dVRK)}
% How action-conditioned WMs act as "Digital Twins" for surgical robotics, replacing expensive physical trials with predictive simulations.

% \subsection{Wireless Resource Planning}
% A multimodal application where the WM predicts both robot movement and channel state information (CSI) to optimize 6G power consumption.

% \subsection{Earth Observation (AnySat)}
% Applying JEPA to planetary-scale data to create a foundation model that is invariant to sensor resolution or modality.

% \subsection{Marine Robotics: The Underwater Frontier}
% An emerging sector where WMs must handle extreme perceptual degradation (turbidity) and high-latency acoustic communication.

% \section{Synthesis: Common Challenges in WMing}
% This section compares how these different domains handle \textbf{Action-Conditioning}, \textbf{Temporal Consistency}, and the \textbf{Uncertainty of Imagination} (e.g., ensemble disagreement).
% \end{revremoveblock}

\section{Lessons from Adjacent Domains and Requirements for Marine Transfer}
Results from terrestrial robotics, autonomous driving, Earth observation, and other physical-AI domains offer valuable transfer evidence but should not be counted as marine validation. Their main contribution is exposing reusable design choices—action conditioning, multi-step latent consistency, multimodal alignment, closed-loop planning, online correction—while marine transfer adds requirements weakly represented in current benchmarks: wave-induced motion, wind/current forcing, intermittent communication, power-limited edge computation, and high consequences of unsafe adaptation. The synthesis therefore assigns each study an evidence level—from conceptual/adjacent-domain evidence through marine simulation, hardware-in-the-loop evaluation, model-scale testing, to full-scale field deployment—clarifying whether a claimed capability has realistic marine validation or remains a research hypothesis.

\section{Review Protocol}
\label{sec:background}

This review examines the architectural distinctions among representation learners, generative simulators, predictive latent models, and action-conditioned WMs, linking them to the maritime platforms, sensing modalities, planning tasks, environmental shifts, and computational constraints evaluated in the literature. It assesses evidence for deployment-time adaptation, including AdaJEPA, with particular attention to marine uncertainty and safety. It also explores applications in adaptive observation, ocean digital twins, and biodiversity monitoring, distinguishing demonstrated capabilities from prospective uses.

\subsection{Scope and Search Strategy}
% \begin{revremoveblock}
% Describe inclusion/exclusion criteria, databases searched, and date ranges.
This review followed a systematic approach to identify, screen, and synthesise literature on WMs for ocean intelligence, retrieved from Scopus, Web of Science, and arXiv using Boolean keyword combinations, restricted to 2019–2026 to capture recent WM advances. Inclusion criteria comprised studies proposing or evaluating WMing frameworks (environment representation and future-state prediction supporting planning or control); non-qualifying studies were excluded. After removing duplicates, title/abstract screening, and full-text review, we included 36 studies. We identified additional literature through secondary sources such as technical blogs and community-curated repositories, subject to the same inclusion and exclusion criteria.
% It maps a rapidly developing, heterogeneous body of journal articles, conference papers, preprints, datasets, and adjacent-domain evidence. 
\begin{longtable}{@{}p{2.5cm}p{12cm}p{2.0cm}@{}}
\caption{Literature search strategy summary.} \label{tab:search} \\
\toprule
\textbf{Database} & \textbf{Search Terms} & \textbf{Records Found} \\
\midrule
\endfirsthead

\midrule
\endhead

\bottomrule
\endlastfoot

Scopus            & ``WM'' & \\
                  &     ``WM'' AND ``JEPA''                    & 03\\
                  & ``WM'' AND ``Review''& 00 \\
Web of Science    & ``WM'' & 04 \\
                  &     ``WM'' AND ``JEPA''                    &     \\
                  & ``WM'' AND ``Review''& 02 \\
arXiv                  &     ``WM'' AND ``JEPA''                    &     \\
                  &     ``WM'' AND ``Cosmos''                    &    03 \\
                  &``Dreamer'' AND ``WM'' AND  (``reinforcement learning'' OR ``latent state'' OR ``RSSM'')& 18\\
Other (blogs, snowballing, preprints)& &06 \\
\midrule
\textbf{Total (after deduplication)} & & \textbf{36} \\

\end{longtable}
\section{Key Findings}
\label{sec:results}

\subsection{Marine WMs Remain Predominantly Pre-Deployment Models}

The reviewed literature indicates that most models are trained offline and remain fixed during evaluation. Direct evidence for long-duration marine deployment, online recalibration, and adaptation to coupled sensor and dynamics shifts remains limited. Consequently, treat strong results from controlled robotics or driving benchmarks as transfer evidence rather than confirmation of robust ocean autonomy.

\subsection{Action Conditioning and Deployment Adaptation Separate WMs from Encoders}

The most operationally relevant methods predict how latent state changes under actions or interventions and expose these predictions to a planner. AdaJEPA adds a further distinction by updating the predictive model from transitions observed during closed-loop MPC~\cite{wang2026adajepa}. This mechanism directly addresses non-stationarity, but current evidence does not establish safe adaptation on marine platforms or ecological timescales.

\subsection{Biodiversity Applications Are Promising but Mostly Prospective}

WMs offer a coherent framework for combining physical, acoustic, visual, molecular, and robotic observations, yet the literature does not currently demonstrate a unified marine biodiversity WM at ecosystem scale. The strongest near-term use case is adaptive observation: a vehicle-level model predicts sensor and environmental evolution, quantifies where new measurements are valuable, and selects safe, energy-aware sampling actions.

%%=============================================
%% 5. DISCUSSION
%%=============================================
\section{Discussion}
\label{sec:discussion}

% \begin{revremoveblock}
% \subsection{Interpretation of Findings}
% Synthesize the findings in the context of the broader field.

% \subsection{Open Challenges}
% \begin{enumerate}
%     \item \textbf{Challenge 1}: Description of challenge and why it is unresolved.
%     \item \textbf{Challenge 2}: Description of challenge.
%     \item \textbf{Challenge 3}: Description of challenge.
% \end{enumerate}

% \subsection{Limitations of This Review}
% Acknowledge limitations such as language bias, publication bias, or scope constraints.
% \end{revremoveblock}

\subsection{Interpretation of Findings}

The central opportunity for ocean engineering lies not in a single model family but in integrating predictive state representation, environmental forcing, action-conditioned dynamics, uncertainty, and planning. JEPAs offer a useful latent-prediction objective, while generative and hybrid models may retain advantages for interpretable observations or scenario visualisation. Deployment adaptation is especially relevant since marine systems cannot assume offline training conditions match real visibility, vehicle response, payload, sea state, or biological context—but this same adaptivity risks updating from sensor faults or safety-critical anomalies. Marine use therefore requires change detection, uncertainty-gated learning, constrained updates, rollback, and independent safety supervision, with adaptation performance reported alongside failure frequency, not just average success.

\subsection{Open Challenges}

\begin{enumerate}
    \item \textbf{Uncertainty and calibration:} separate predictable dynamics, irreducible stochasticity, epistemic uncertainty, and sensor failure.
    \item \textbf{Multi-scale dynamics:} reconcile sub-second control, tidal and weather evolution, seasonal ecology, and climate-scale change in a common hierarchy.
    \item \textbf{Safe online adaptation:} prevent corrupted transitions, catastrophic drift, or short-term specialisation from degrading navigation and monitoring.
    \item \textbf{Embedded operation:} validate inference and adaptation latency, energy, memory, bandwidth, and thermal load on realistic USV, AUV, and glider computers.
    \item \textbf{Marine benchmarks:} provide synchronised multimodal datasets and repeatable shifts involving visibility, waves, currents, payload, sensor fouling, geography, and traffic.
    \item \textbf{Ecological validity:} preserve rare species and anomalous events rather than treating them as noise, and quantify sampling bias and forecast calibration.
\end{enumerate}

\subsection{Limitations of This Review}

This review spans rapidly evolving ML and ocean-engineering literatures, including preprints whose findings may change after peer review. Terminology is inconsistent across communities, with some WM-labeled works lacking action-conditioned dynamics or planning, and the breadth of adjacent-domain evidence risks overstating marine transferability. These limitations are mitigated through explicit architectural definitions, evidence-level coding, and reproducible search reporting.

%%==============================================
%% 6. FUTURE DIRECTIONS
%%==============================================

\section{Future Directions: Beyond Robotics}
\label{sec:future}
\subsection{Marine WMs for Biodiversity Intelligence}
\begin{figure}
    \centering
    \includegraphics[width=5in, height=2.6in]{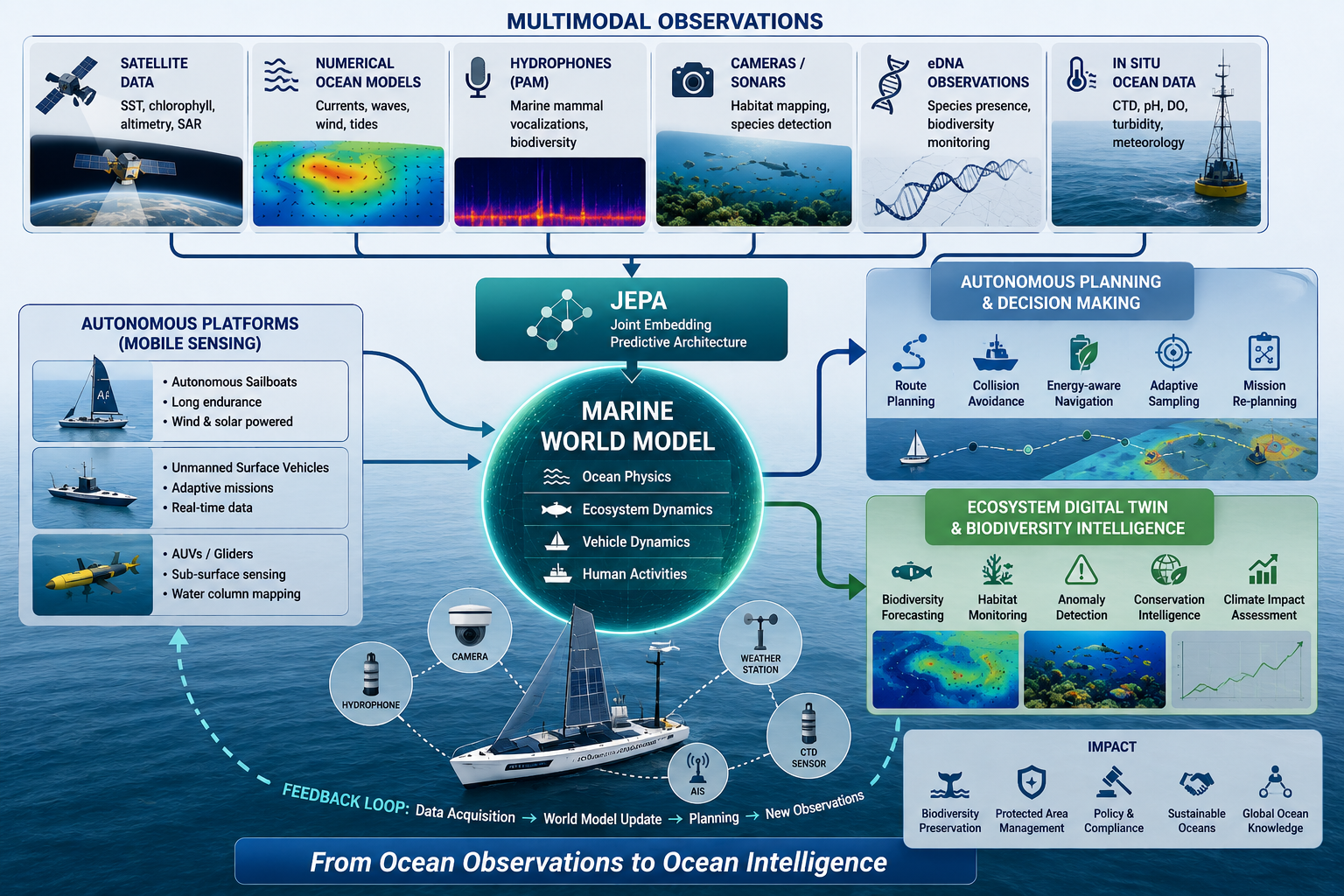}
    \caption{Biodiversity WM for Adaptive Ocean Monitoring}
    \label{fig:biodiversity_wm}
\end{figure}

Traditional biodiversity monitoring relies on sparse field surveys and retrospective analysis, offering fragmented views of dynamic ecological processes. WMs offer a different paradigm~\cite{ding2025understanding,lecun2022path}: a Biodiversity WM learns a latent representation of ecosystem dynamics that can predict future states~\cite{ha2018world,hafner2019dream,hafner2023mastering}. Heterogeneous observations—satellites, autonomous vehicles, hydrophones, eDNA—are projected into a shared latent space, enabling forecasts of species distributions and biodiversity hotspots. As shown in Figure~\ref{fig:biodiversity_wm}, the WM lets autonomous platforms transition from passive sensing to active ecological exploration.

\subsection{Scalable Biodiversity and Ecosystem Modeling}
% \begin{revremoveblock}
% An emerging frontier for WMs is global \textbf{Biodiversity Modeling}, where current ecological models often struggle with sampling bias and noisy real-world datasets. Applying the JEPA framework enables WMs that: (i) \textbf{filter perceptual noise} by capturing predictable biological dynamics (e.g., migration patterns, species interactions) while ignoring task-irrelevant fluctuations like weather in satellite imagery~\cite{assran2025v}; (ii) build \textbf{digital twins of ecosystems}—large-scale forest or marine habitat models that can "imagine" climate change or deforestation impacts decades in advance~\cite{islam2026fair}; and (iii) enable \textbf{multimodal wildlife monitoring} by integrating wireless acoustic sensors with remote sensing data to track species occupancy and health in real time, even in remote, data-scarce regions~\revremove{[citation required]}.
% \end{revremoveblock}

Global biodiversity modelling remains a prospective extension of world-model research, not an established JEPA capability. A defensible near-term goal is learning predictive ecological representations that integrate physical ocean variables with acoustic, visual, molecular, and occurrence data while explicitly modelling sampling bias, delayed measurements, and uncertainty. Such representations could support ecosystem digital twins and multimodal monitoring, but decadal forecasting claims require validation against mechanistic models, calibrated uncertainty, and temporally independent observations—without discarding rare species, abrupt transitions, or anthropogenic disturbances as task-irrelevant noise.

A key capability enabled by Biodiversity WMs is information-driven exploration: rather than following predefined transects, autonomous sailboats can select trajectories maximising expected reduction in ecosystem uncertainty, formulated as an information gain optimisation problem where the WM predicts both future states and the value of prospective observations—transforming marine robots from passive platforms into adaptive instruments capable of autonomously discovering biodiversity hotspots and ecological anomalies~\cite{reygondeau2026aquax}.

\subsection{Ecological State Representation}

A fundamental challenge in biodiversity monitoring is that ecosystems are only partially observable: measurements from satellites, hydrophones, camera systems, eDNA, oceanographic sensors, and autonomous robotic platforms provide fragmented views of complex ecological processes evolving across multiple spatial and temporal scales. Traditional monitoring estimates individual variables (species abundance, habitat extent, environmental parameters), but effective forecasting requires a more holistic representation capturing whole-ecosystem dynamics while preserving real correlations and interdependence~\cite{etheve2026diel}.

WMs offer a promising framework. Rather than directly predicting future observations, a WM learns a latent representation capturing the predictable components of ecosystem dynamics. Formally, let

\begin{equation}
x_t = {x_t^{sat},x_t^{ac},x_t^{cam},x_t^{eDNA},x_t^{env}}
\end{equation}

denote multimodal observations at time $t$ (satellite, acoustic, imaging, eDNA, and environmental/oceanographic variables, respectively), which an encoder $E(\cdot)$ maps into a latent ecological state

\begin{equation}
z_t = E(x_t),
\end{equation}

a compact ecosystem description. Unlike conventional ecological indicators, this latent state is not explicitly constrained to predefined variables; instead, it emerges from SSL and can encode species richness, trophic interactions, habitat condition, ecosystem resilience, biodiversity hotspots, and anthropogenic pressures.

The ecosystem's temporal evolution can then be modelled via a latent dynamics function

\begin{equation}
z_{t+1}=f(z_t,u_t,e_t),
\end{equation}

where $u_t$ represents management actions or robotic interventions and $e_t$ represents external environmental forcing (temperature, currents, winds, storms, climate perturbations), enabling prediction of future states without simulating every observable variable.

In a JEPA-based framework, the objective is to predict informative latent representations of future observations rather than reconstruct full sensory input~\cite{lecun2022path}—naturally filtering measurement noise to focus on ecological dynamics consistent across sensing modalities and conditions.

The AdaJEPA analogy holds most strongly for the robotic sensing loop, where new vehicle–environment transitions can recalibrate short-horizon predictions. It is weaker for ecosystem-state learning, given delayed eDNA sampling, intermittent acoustic observations, and biological processes unfolding far slower than an MPC step. Ecological deployment therefore requires multi-rate state estimation, delayed-observation updates, seasonal memory, and safeguards against adapting away rare but meaningful events.

The resulting Ecological WM serves as the predictive core of a biodiversity digital twin, continuously assimilating satellite, fixed-infrastructure, and autonomous-vehicle observations to maintain an evolving ecosystem-state representation and generate forecasts—supporting species distribution forecasting, habitat degradation assessment, and harmful algal bloom prediction. This is especially relevant for marine environments, where observations are sparse and costly. Long-endurance sailboats and USVs can act as mobile sensing agents that continuously update the latent ecological state through adaptive sampling, exploiting the WM's predictive capacity to identify regions where additional measurements maximise information gain:
% \begin{revremoveblock}
% \begin{equation}
% a_t^* =
% \arg\max_{a_t}
% I(z_{t+1};o_{t+1})
% \end{equation}
% \end{revremoveblock}

\begin{equation}
a_t^*=\arg\max_{a_t\in\mathcal{A}}
I\!\left(z_{t+1};o_{t+1}\mid\mathcal{D}_t,a_t\right)
-\lambda_E C_E(a_t)-\lambda_R C_R(a_t),
\end{equation}
where $\mathcal{D}_t$ is the information available at planning time, $C_E$ is an energy cost, $C_R$ is a safety or mission-risk cost, and $\lambda_E$ and $\lambda_R$ control their trade-off with expected information gain.

where $z_{t+1}$ is the future ecosystem state and $o_{t+1}$ the future observation. Biodiversity monitoring thus evolves from passive observation toward an active perception framework in which autonomous agents continuously refine and validate ecosystem-scale predictive representations.

\subsection{Challenges in Ecological Imagination}
The transition to ecological WMing is not without difficulty. Marine ecosystems combine partially observed, nonlinear, stochastic, and externally forced processes across widely separated temporal and spatial scales. Future research must focus on:
\begin{enumerate}
    \item \textbf{Uncertainty Quantification:} Distinguishing between what a model can predict and what is fundamentally stochastic in nature.
    \item \textbf{Global-to-Local Scaling:} Developing JEPA architectures that remain accurate across different pixel resolutions, from individual camera traps to planetary-scale satellite arrays.
\end{enumerate}

%\subsection{Edge Data Acquisition: USVs as Mobile Sensory Hubs}
\subsection{World-Model-Driven Ocean Exploration}
Unlike terrestrial robotics, ocean science has a long-standing culture of open data sharing and increasingly embraces FAIR principles~\cite{wilkinson2016fair}, which matters for Marine WMs that learn coherent representations from heterogeneous, independently collected data. Decades of satellite, platform, and repository observations span physical, chemical, biological, and geological processes~\cite{goos2030, revelard2022ocean}. To move WMs from simulation to conservation tools, we propose integrating \textbf{Unmanned Surface Vehicles (USVs)} as high-endurance data nodes. Wind- and solar-powered USVs, such as \href{https://www.saildrone.com}{Saildrone platforms}, operate for extended periods, providing atmospheric and oceanographic data, while \textbf{high-stability catamarans} support hyperspectral and LiDAR imaging to map coral reefs and seagrass beds. These multimodal streams feed the \textbf{JEPA latent space}, letting the WM predict biological dynamics like fish movement even when sensors are obscured by turbidity.

\subsection{Proactive Conservation in Protected Areas}
By deploying USV fleets in Marine Protected Areas (MPAs), the WM moves beyond passive monitoring to proactive \textbf{Conservation Intelligence}:
\begin{enumerate}
    \item \textbf{Autonomous Patrols:} USVs can detect illegal fishing or shipping incursions through real-time AIS and machine learning-based target recognition.
    \item \textbf{Habitat Digital Twins:} Combining USV-derived high-resolution bathymetry and biological data enables the creation of "Living Digital Twins" that simulate the impact of rising sea temperatures or pollution events on local biodiversity.
\end{enumerate}

\subsection{From Data Acquisition to Active Guidance: JEPA-Driven Path Planning}
While traditional drones follow pre-defined waypoints, integrating a WM transforms the USV into an \textbf{adaptive explorer}, shifting from passive sensing to \textit{active, goal-oriented navigation} via JEPA's predictive capabilities. For \textbf{obstacle avoidance}, rather than reactive manoeuvres, a JEPA-based WM predicts \textit{temporal dynamics}~\cite{nam2026causal} of the vehicle and moving obstacles, simulating future latent states to identify a collision-free trajectory accounting for hydrodynamics and current drift. For \textbf{gradient-based exploration}, water/air-quality sensors (pH, dissolved oxygen, turbidity, CO2, VOCs) act as a "scientific landscape," where the agent's "imagination" predicts likely anomaly locations—e.g., guiding the USV toward an illegal discharge point or biological hotspot, optimising information gain per battery cycle. Finally, for \textbf{energy-aware informative planning}, the WM predicts trade-offs between wind-energy harvesting and scientific value, learning a path maximising data density while preserving power for long-term MPA conservation missions.
%\section{Bridging Technology and Policy}
\section{From Marine WMs to Ocean Intelligence}
The previous sections traced the evolution of World Models (WMs) from latent predictive representations in robotics to emerging approaches for modelling complex ecological systems. Beyond individual-agent decision-making, WMs could ultimately enable ecosystem-scale intelligence by integrating autonomous marine platforms, sensing infrastructures, numerical ocean models, and biodiversity monitoring into shared predictive representations, with Marine WMs serving as a cognitive layer for next-generation Ocean Intelligence systems.

\subsection{Open Ocean Data as the Foundation of Marine WMs}

The remarkable progress of Foundation Models in NLP and computer vision has largely been driven by massive open datasets~\cite{bommasani2021opportunities}: LLMs learn from trillions of internet-collected words, while Vision FMs train on millions of images and videos~\cite{ding2025understanding}. Likewise, future Marine WMs will depend fundamentally on large-scale, heterogeneous, continuously updated ocean observations.

Unlike terrestrial robotics, the ocean science community has a long-standing culture of open data sharing through international observing programs, increasingly embracing FAIR (Findable, Accessible, Interoperable, Reusable) principles~\cite{wilkinson2016fair}, which have transformed observing infrastructures into interoperable ecosystems supporting large-scale multimodal integration—particularly significant since Marine WMs depend on learning coherent latent representations from heterogeneous, independently collected observations. Over recent decades, satellites, autonomous platforms, fixed observatories, biodiversity repositories, and numerical prediction systems have generated an unprecedented volume of publicly available observations spanning physical, chemical, biological, acoustic, and geological processes~\cite{goos2030, revelard2022ocean}.

\begin{figure}[h!]
    \centering
    \includegraphics[width=5in, height=2.5in]{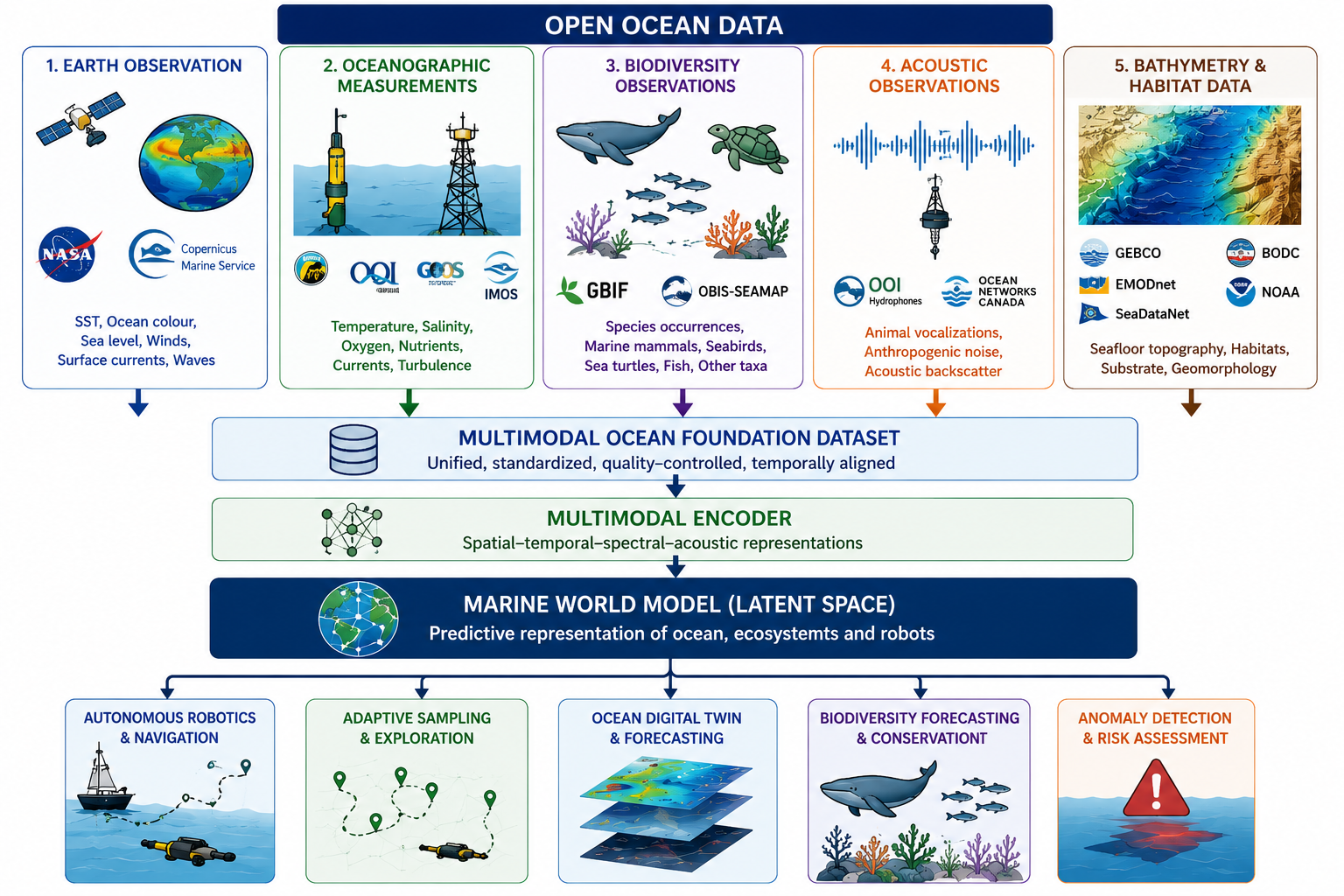}
    \caption{Conceptual overview of the data-centric paradigm for Marine WMs. Decades of open, FAIR-compliant ocean observations—satellite remote sensing, oceanographic monitoring, biodiversity repositories, and acoustic observatories—provide the multimodal training data for learning predictive marine representations. Integrating these sources into a common latent space enables future-state prediction, adaptive robotic exploration, digital twins, and biodiversity forecasting, laying the foundation for next-generation Ocean Foundation Models supporting intelligent ocean observation at global scale.}
    \label{fig:open_ocean_data}
\end{figure}

As shown in Figure~\ref{fig:open_ocean_data}, this diversity complements the multimodal nature of Marine WMs across five categories. \textbf{Earth observation datasets} (NASA EarthData~\cite{nasaearthdata}, Copernicus Marine Service~\cite{copernicusmarine}) provide synoptic observations of SST, ocean colour, chlorophyll, sea level, winds, waves, and currents—describing large-scale physical context. \textbf{Oceanographic observing systems} (GOOS, Argo, OOI, IMOS) continuously monitor temperature, salinity, oxygen, and currents, characterising water-column evolution~\cite{argo2001argo}. \textbf{Marine biodiversity repositories} (GBIF~\cite{telenius2011biodiversity}, OBIS-SEAMAP~\cite{halpin2009obis}) provide species occurrence records enabling models to learn ecosystem structure and biodiversity hotspots. \textbf{Passive acoustic networks} (Ocean Networks Canada, OOI hydrophones) contribute recordings of marine vocalisations and anthropogenic noise~\cite{fleishman2023ecological}, informing ecosystem monitoring~\cite{ragland2022overview}. Finally, \textbf{bathymetric and habitat datasets}~\cite{bloch2025overview} (GEBCO~\cite{weatherall2025gebco_2025}, EMODnet~\cite{emodnet2024}, SeaDataNet~\cite{schaap2026seadatanet}, BODC, NOAA) describe seabed morphology and benthic habitats, providing spatial context for navigation and ecosystem modelling.

By integrating heterogeneous sources into a common latent representation, Marine WMs can jointly capture ocean physics, ecosystem dynamics, and human activities, supporting adaptive sampling, navigation, and conservation intelligence. FAIR observations~\cite{wilkinson2016fair} enable a shift to continuously evolving systems, updated through continual learning and data assimilation~\cite{parisi2019continual,hafner2025mastering}. International ocean observing systems thus provide both the training data and infrastructure for future Ocean FMs at global scale.

Table~\ref{tab:open_ocean_data} summarises the principal categories of open ocean datasets, the environmental information they provide, and their potential contribution to Marine WMs.

\begin{longtable}{p{2.5cm} p{3.0cm} p{4.0cm} p{4.5cm} p{0.5cm}} 
\caption{Representative open ocean data sources and their potential contribution to Marine WMs.}
\label{tab:open_ocean_data}
\renewcommand{\arraystretch}{1.2}
\\
\textbf{Open Ocean Data Source} &
\textbf{Primary Data} &
\textbf{Marine WM Component} &
\textbf{Example Applications} &
\textbf{Reference} \\
\midrule

NASA EarthData &
Sea surface temperature, ocean colour, sea level, winds &
Large-scale ocean physics and environmental forcing &
Ocean forecasting, mission planning, environmental awareness &
\cite{nasaearthdata} \\

Copernicus Marine Service &
Physical and biogeochemical forecasts &
Ocean dynamics and environmental prediction &
Adaptive navigation, long-term mission planning &
\cite{copernicusmarine} \\

GOOS &
Global ocean observations &
Multiscale ocean state representation &
Global ocean monitoring, data assimilation &
\cite{goos2030} \\

Argo Program &
Temperature, salinity, oxygen profiles &
Water-column dynamics &
Ocean state estimation, latent representation learning &
\cite{argo2001argo} \\

Ocean Observatories Initiative (OOI) &
Moorings, cabled arrays, gliders, hydrophones &
Continuous multimodal observations &
Marine ecosystem monitoring, acoustic intelligence &
\cite{smith2018ooi}\\

Integrated Marine Observing System (IMOS) &
Physical, biological, acoustic and biogeochemical observations &
Multimodal environmental representation &
Regional ecosystem monitoring &
\cite{hill2010imos} \\

GBIF &
Species occurrence records &
Biodiversity representation &
Species distribution modelling, ecological forecasting &
\cite{telenius2011biodiversity} \\

OBIS-SEAMAP &
Marine mammals, seabirds and sea turtles &
Marine biodiversity dynamics &
Wildlife monitoring, conservation &
\cite{halpin2009obis} \\

Ocean Networks Canada (ONC) &
Continuous hydrophone and observatory data &
Marine acoustic environment &
Passive acoustic monitoring, marine mammal detection &
\cite{ragland2022overview} \\

GEBCO &
Global bathymetry &
Seafloor geometry &
Navigation, habitat characterisation &
\cite{weatherall2025gebco_2025} \\

EMODnet &
Bathymetry, biology, habitats, geology, chemistry &
Marine habitat representation &
Marine spatial planning, ecosystem modelling &
\cite{emodnet2024} \\

SeaDataNet &
Marine environmental observations &
Historical environmental knowledge &
Long-term environmental analysis &
\cite{schaap2026seadatanet} \\

NOAA NCEI &
Oceanographic, fisheries and climate archives &
Long-term ocean state and climate variability &
Climate studies, environmental forecasting &
\cite{wod2025} \\

British Oceanographic Data Centre (BODC) &
Cruises, moorings, oceanographic observations &
Long-term ocean observations &
Ocean process modelling and validation &
\cite{bodc2026}\\

\bottomrule
\end{longtable}

\subsection{Marine WMs as the Core of Ocean Digital Twins}
As shown in Figure~\ref{fig:marine_wm}, the Marine WM serves as a common predictive layer linking heterogeneous observations, robotic platforms, and ecosystem-scale forecasting—extending digital twins beyond physical ocean variables to integrated ecological representations. Digital Twin technologies are a promising framework for modelling complex environmental systems, and current ocean digital twins already combine simulations, remote sensing, and observational data; however, most remain heavily physics-based and struggle to integrate heterogeneous biological observations across varying spatial and temporal scales.

Marine WMs offer a complementary paradigm: by learning predictive latent representations from multimodal observations, they continuously assimilate data from satellites, oceanographic sensors, hydrophones, eDNA, and autonomous platforms—generating plausible future ecosystem trajectories rather than just reconstructing current state, enabling anticipatory rather than reactive management. In this way, the WM functions as the predictive engine of an Ocean Digital Twin, continuously updated by incoming observations and capable of forecasting both physical and ecological dynamics.

\subsection{Distributed Ocean Sensing through Autonomous Platforms}
A key limitation in biodiversity monitoring is observation sparsity, as vast ocean regions remain undersampled due to the high cost of research vessels and fixed infrastructure. Long-endurance autonomous platforms—robotic sailboats, USVs, gliders, and AUVs—offer a scalable alternative, continuously collecting multimodal acoustic, visual, and oceanographic data. When integrated with a Marine WM, these platforms move beyond fixed survey routes: using predictive representations, they can target high-uncertainty regions, maximise information gain, and adapt sampling in real time—turning ocean observation into an adaptive, self-optimising sensing ecosystem.

\subsection{Biodiversity Forecasting and Conservation Intelligence}
Marine WMs hold strong promise for biodiversity forecasting, moving conservation beyond retrospective, historical analysis toward proactive prediction. By combining ecological observations with predictive latent representations, Biodiversity WMs could forecast ecosystem states under varying environmental and human pressures—aiding early detection of invasive species, habitat degradation, biodiversity hotspots, and climate-driven change. Paired with autonomous sensing platforms, this enables a new paradigm of Conservation Intelligence, where systems not only monitor conditions but anticipate ecological risks and recommend adaptive interventions.

\subsection{Towards Ocean Intelligence}
The convergence of WMs, autonomous robotics, digital twins, and biodiversity monitoring points toward a broader paradigm—Ocean Intelligence—where predictive representations act as a shared computational substrate linking observation, autonomous decision-making, and ecosystem forecasting. Future systems could continuously fuse satellite, robotic, citizen science, eDNA, and simulation data into an evolving marine representation that reasons about possible futures and informs conservation policy, rather than simply describing present conditions. In this sense, Marine WMs mark a shift from data-driven observation to predictive ocean understanding, laying the groundwork for intelligent, adaptive ocean systems—with Ocean Foundation Models trained on decades of satellite, AIS, acoustic, and biodiversity data as a natural next step.

\begin{longtable}{p{3.0cm}p{7.0cm}p{7.0cm}}
\caption{Candidate world-model families for marine applications and the evidence required before deployment.} \label{tab:marine_application_matrix} \\
\toprule
\textbf{Model family} & \textbf{Candidate marine use} & \textbf{Principal validation gap} \\
\midrule
\endfirsthead

\midrule
\endhead

\bottomrule
\endlastfoot

Video or multimodal JEPA & Perception-aware navigation, docking, obstacle prediction, and short-horizon environmental forecasting & Demonstrate action-conditioned temporal prediction under turbidity, glare, wave-induced motion, sensor displacement, and moving traffic. \\
Generative video WM & Operator visualisation, scenario generation, and simulation-assisted training & Establish physical consistency, uncertainty calibration, and real-time operation with hydrodynamic and navigation constraints. \\
Trajectory- or action-conditioned latent model & Route planning, collision avoidance, adaptive transects, and energy-aware missions & Validate on unstructured marine geometry with currents, winds, bathymetry, COLREG constraints, and long planning horizons. \\
AdaJEPA-style adaptive model & Recalibration to changing sensors, vehicle response, currents, payload, or local observation conditions & Add uncertainty-gated updates, corruption rejection, rollback, embedded-compute evaluation, and model-scale or full-scale marine trials. \\
Ecological latent model & Biodiversity state estimation, information-driven sampling, and digital-twin assimilation & Support delayed and multi-rate observations, rare-event preservation, ecological calibration, and validation against independent time periods and process-based models. \\

\end{longtable}

% \revremove{Table~\ref{tab:marine_challenges} summarizes the principal challenges faced by autonomous marine systems and highlights how WMs provide a unified framework for addressing many of the limitations of conventional perception-driven autonomy architectures..}

Table~\ref{tab:marine_application_matrix} separates candidate applications from the marine evidence still required, avoiding the implication that adjacent-domain results already demonstrate deployment readiness.

%\subsection{Edge-Inference and Latent Data Acquisition}
%We evaluate the role of \textbf{autonomous sailboat drones} as mobile sensory nodes. By utilizing JEPA's predictive capabilities, these vehicles can operate in \textit{disconnection-tolerant} modes, transmitting only latent state summaries over low-bandwidth acoustic or satellite links.

%\subsection{Proactive Enforcement and the Legal Frontier}
%The transition from monitoring to \textit{enforcement} requires WMs that can generate "explainable" predictions for maritime authorities. We discuss the alignment of these AI systems with the \textbf{International Maritime Organization (IMO)}'s four levels of autonomy and the evidentiary requirements for prosecuting environmental crimes in MPAs.

%%==============================================
%% 7. CONCLUSION
%%==============================================
\section{Conclusion}
\label{sec:conclusion}
Marine world models (WMs) remain an emerging technology: most are trained offline and stay static during deployment, so strong benchmark results don't yet confirm robustness for long-duration ocean use or shifting conditions. True WMs are distinguished from simple encoders by predicting how latent states change under actions to guide planning—methods like AdaJEPA go further by updating from closed-loop experience, though safe adaptation on real marine platforms remains unproven. While WMs show promise for integrating diverse biodiversity data, no unified ecosystem-scale model yet exists; the most realistic near-term use is adaptive observation, guiding efficient, energy-aware sampling.
WMs offer a promising framework for ocean engineering, integrating state estimation, prediction, and decision-making under partial observability. However, the term is often used too loosely—representation learning, dynamics modelling, action conditioning, and planning should be treated as distinct capabilities, not assumed together. AdaJEPA illustrates progress by adapting from closed-loop MPC transitions to recover performance under simulated shifts~\cite{wang2026adajepa}, though marine validation is still lacking. The most promising path forward is a deployment-adaptive marine WM combining physics-informed dynamics and guarded online correction, tested progressively toward field trials, with biodiversity uses starting from adaptive observation rather than full digital twins.

%%==============================================
%% ACKNOWLEDGEMENTS
%%==============================================
\section*{Acknowledgements}
 The authors gratefully thank Mr Arturo Argentieri from CNR-ISASI, Italy, for his technical contribution to the multi-GPU computing facilities. 

%%==============================================
%% FUNDING
%%==============================================
\section*{Funding}
This work was supported in part by the ROMEO project funded under the Sustainable Blue Economy Partnership Call 2024, CUP B83C24009260005, and AI4Nature CUP B29H26000110007.

%%==============================================
%% DECLARATION OF INTERESTS
%%==============================================
\section*{Declaration of Competing Interests}
The authors declare that they have no known competing financial interests or personal
relationships that could have appeared to influence the work reported in this paper.

%%==============================================
%% AUTHOR CONTRIBUTIONS (CRediT)
%%==============================================
\section*{Author Contributions}

\textbf{BD}: Conceptualization, Data Curation, Methodology, Writing -- Original Draft.
\textbf{EWC}: Supervision, Formal Analysis.
\textbf{LMG}:Writing -- Review \& Editing;
\textbf{PS}:  Funding Acquisition;
\textbf{CD}: Supervision, Funding Acquisition;

\section*{Declaration of Generative AI and AI-Assisted Technologies in the Manuscript Preparation Process}

During the preparation of this work, the authors used OpenAI and Grammarly to assist with manuscript organisation, critical figure editing, and language revision. The authors reviewed and edited the resulting material and take full responsibility for the manuscript's content.

%%==============================================
%% REFERENCES
%% Uses natbib — compile with BibTeX
%%==============================================
\bibliographystyle{elsarticle-harv}  %% author-year style required by the journal
\bibliography{cas-refs}            %% references.bib file

\end{document}